# AgriBot: Agriculture-Specific Question Answer System

Naman Jain
Discipline of Computer Science
Indian Institute of Technology Gandhinagar
Gandhinagar, India
naman.j@iitgn.ac.in

Pranjali Jain
Discipline of Computer Science
Indian Institute of Technology Gandhinagar
Gandhinagar, India
pranjali.jain@iitgn.ac.in

Pratik Kayal
Discipline of Computer Science
Indian Institute of Technology Gandhinagar
Gandhinagar, India
pratik.kayal@iitgn.ac.in

P Jayakrishna Sahit
Discipline of Computer Science
Indian Institute of Technology Gandhinagar
Gandhinagar, India
jayakrishna.sahit@iitgn.ac.in

Soham Pachpande
Discipline of Computer Science
Indian Institute of Technology Gandhinagar
Gandhinagar, India
pachpande.soham@iitgn.ac.in

Jayesh Choudhari
Discipline of Computer Science
Indian Institute of Technology Gandhinagar
Gandhinagar, India
choudhari.jayesh@iitgn.ac.in

Mayank Singh
Discipline of Computer Science
Indian Institute of Technology Gandhinagar
Gandhinagar, India
singh.mayank@iitgn.ac.in

*Abstract* — **India is an agro-based economy and proper information about agricultural practices is the key to optimal agricultural growth and output. In order to answer the queries of the farmer, we have build an agricultural chatbot based on the dataset from Kisan Call Center. This system is robust enough to answer queries related to weather, market rates, plant protection and government schemes. This system is available 24*7, can be accessed through any electronic device and the information is delivered with the ease of understanding. The system is based on a sentence embedding model which gives an accuracy of 56%. After eliminating synonyms and incorporating entity extraction, the accuracy jumps to 86%. With such a system, farmers can progress towards easier information about farming related practices and hence a better agricultural output. The job of the Call Center workforce would be made easier and the hard work of various such workers can be redirected to a better goal.**

## I. Introduction

In India, agriculture plays an important role in the economic development by contributing about 16% to the overall GDP and accounting for employment of approximately 52% of the Indian population[12]. According to the Farmers' portal[12], rapid growth in agriculture is essential not only for self-reliance but also to earn valuable foreign exchange. However, most farmers do not have access to authentic information about the latest farming practices and trends. One of the reasons is that the people involved in the occupation of farming are comparatively slow adopters of latest technology. Traditionally, field officers visit the farmlands and provide training, advice, and support to the farmers. Many of the rural villages lack the ease of accessibility which results in the wastage of time and money spent on obtaining information or contacting officials. Hence, farmers are often unable to obtain agricultural information which can help them in taking better decisions related to the crops that they cultivate. This leads to reduced crop yield, increased wastage of valuable labor, and market inefficiency. These reasons add up to severely impact a farmer's earnings, time and opportunities to increase the crop yield. In recent years, the use of Information Technology (IT) in agriculture extension has grown. According to TRAI[13], there were 647 million urban mobile subscribers while 519 million rural subscribers as of Aug 2018 and The Economic Times predicted that by 2020, 50% of internet users will be from the rural sector. This data shows that mobile connectivity is seeing an exponential growth aiding to the promotion of agricultural information by IT services. The Government faces difficulties in spreading vital information related to farming. Also, the problems worsens due to the spread of misinformation. These problems are present due to the vast language diversity and lack of confidence of the rural population on modern technologies. In such a scenario, usage of mobile devices to spread agriculture related information appears to be a promising solution.

## II. Related Work

To address the above defined problem, Government has initiated various agriculture related IT services which provide access to a central knowledge bank. Most prominent services are mentioned below:

### Farmer's Portal

Farmer's Portal makes use of the Internet as a tool to make knowledge accessible. Farming related information is available on this website but is mainly presented in English and Hindi. However, one of the significant challenges faced by this service is that most farmers are not literate to operate computers properly. According to a survey by Times of India[14], only 6% of rural households owned a computer, and only 18% of the rural youth knew how to operate one. This survey explains the fact that a website is not a feasible option to spread agriculture related information to farmers because of the lack of computer training.

*AgriApp*

AgriApp is one of the most popular apps among farmers. It has a rating of 4.3 out of 5 on Google Play store. This portal brings information about farming resources and government services through an online mobile application to the farmers. It also provides a chat option for farmers which enables them to chat with an agricultural expert using this app efficiently. However, AgriApp is a knowledge bank wherein the user has to search for a particular piece of information manually and if the user opts to chat with the application operator instead of searching manually, the user has to wait for a significant period of time for a response from the operator.

*Kisan Call Center (KCC)*

KCC is a helpline service for farmers to clarify their queries over the phone. Since the service facilitates a telephonic conversation, this service is able to cater to the need of the farmers on an individual basis as the information is provided in their native language and relevant to their location. Also, the farmers get valuable information related to new farming practices. This service reduces the difficulty of the farmer to ask for help related to latest agricultural practices which also helps in building the trust of the rural class on the Government. However, KCC services are only available from 6 AM to 10 PM, and skilled labor with good knowledge of agricultural practices is required to operate the Call Center. Also, it is observed that with time, queries to KCC have increased exponentially due to increase in awareness among farmers as well as technology adoption. This has the potential to generate the need to set up new call centers which will require massive cost along with training the human resource.

According to the analysis of Kisan Call Center data, about 1.36 million calls were made to Kisan Call Centers in 2017 which increased to about 1.72 million calls in 2018. This shows a 21% increase in calls from 2017 to 2018. In Maharashtra itself, 92% of calls were redundant in the year 2017 and for all the states, only 5% unique new queries were made in 2018 compared to 2017. The number of questions are increasing gradually, and soon these call centers may not be able to efficiently answer all these queries on time, plus most of the queries are redundant. Hence, a scalable solution is needed to accommodate the increase in the number of queries in a better way. We use the power of Artificial Intelligence(AI) to build a solution to this problem.

There exists a good number of Q&A models which deal with a similar problem. Authors in [15] use a knowledge graph based method, where the knowledge graph is built upon the data and questions are answered using the knowledge graph. Another work [16] is a comprehension based question answering system. In such systems, for every question the system generates an answer based on the knowledge gathered by understanding the comprehensions. However, these methods cannot be used to solve our problem because neither our data is properly formatted in a comprehension nor the facts can be extracted to form a knowledge graph. Another way to approach the problem is using Question-Answer pair hashing. However, it must be modified to fit our needs because many semantically similar questions have different answers. Hence, we come up with a practical methodology to implement an agriculture specific AI system.

In a nutshell, right information is crucial for social and economic activities which fuel the development process of a nation. In order to achieve it, we require a decision support solution as simple as a messaging app which makes use of Internet to ease accessibility and automate the process of the conversation with an operator to avoid redundancy. Also, the system should integrate features like real-time outputs, farmer-friendly interface, information delivery in multiple languages, and cost-effectiveness for both the farmer and the operating authority. Such a solution can potentially bridge the information gap for the farmers to facilitate building a productive market.

### III. DATA COLLECTION

We collected our data from https://data.gov.in[9][10]. The collection of each file requires entry of the user's name and email ID. Since we are collecting data for all states of India for the past five years, we automated the whole process through a JavaScript program which downloads and store files as Comma Separated Values (CSV).

For each state, we retrieve district wise CSV. Each file contains the query ID, the query, query-type, query creation time, state name, district name, season and the answer to a given query. Table I shows the total number of queries, namely question-answer pairs in the data. Also, it explicitly shows the number of queries in the years 2017 and 2018.

The data we obtained is not properly formatted and machine readable because it is a summary of the telephonic conversation of a KCC employee with a farmer which has been noted down by the KCC personnel to maintain records.

One of the most critical aspects of the data is that it is multilingual. We observe that some words in the data were written in the native language of a particular state and in some cases the entire data entry had been written in the native language. In addition to that, the data entries do not have proper grammar, spelling or punctuation. Another important aspect about the data is the ambiguity in the responses to the queries. Most of the answers do not completely describe the information asked in the question. Also, a large number of answers related to fertilizer names, or quantities of fertilizer, pesticide or water to be used for the plants are just described in numerical figures. We also noticed that the answer to the same question can be different in different states. Some of the answers to a query vary within the same district plus the answer to a question also depends on the season in which it has been asked.

| Year | Pairs |
|---|---|
| Total number of Pairs (2013-2018) | 99,568,354 |
| Year 2017 | 13,613,465 |
| Year 2018 (Till October) | 17,265,241 |

TABLE I
NUMBER OF QUESTION-ANSWER PAIRS PER YEAR

## IV. DATA ANALYSIS

To understand our dataset, we explored the features we are already presented with, namely, the state names from where each query was asked, the season and query type. Based on this information we can derive the statistics as mentioned in the following tables. We got the statistics related to the number of queries per state (Fig. 2), the number of queries per crop (Table III) as well as crop type (Table II), the number of queries per season (Table VI) and the distribution of queries based on sectors (Fig. 1) as well as query types (Table V).

The data analysis gives a good picture of the agricultural landscape of India regarding which crops are popular in which state, what kind of queries are most commonly asked, and the different sectors the queries are related. For instance, the maximum number of queries asked were related to cereals, specially paddy. Also, the maximum number of queries were asked from the state of Uttar Pradesh. All of these statistics turned out to be factually accurate.

| Query (crop type) | percentage |
|---|---|
| cereals | 31.1% |
| vegetables | 16.6% |
| pulses | 9.5% |
| fruits | 7.8% |
| oilseeds | 7.6% |
| fiber crops | 7.4% |
| millets | 6.5% |
| animal | 3.9% |

TABLE II
QUERY DISTRIBUTION AMONG CROP TYPE

| Query (crop name) | percentage |
|---|---|
| paddy | 30.0% |
| wheat | 20.7% |
| cotton | 12.2% |
| pearl millet | 6.7% |
| sugarcane | 6.0% |
| bovine | 5.7% |
| groundnut | 5.5% |
| black gram | 5.0% |
| bengal gram | 4.1% |
| green gram | 4.1% |

TABLE III
QUERY DISTRIBUTION AMONG CROP NAMES

The number of queries asked during each season is shown in Table VI. This distribution indicates that the number of farmers grows crops during Kharif season is most as compared to other seasons and hence the number of queries for this season are 54%.

Another thing we noticed from the dataset is the number of times each question is repeated. From 2017 to 2018 there was a 5% increase in the number of unique questions. Also, we noticed is that the number of queries related to weather is about 64.4% of the total number of queries asked. For such queries, our model deals differently and quite easily by integrating a weather API.

| Season | Percentage |
|---|---|
| kharif | 56.4% |
| rabi | 28.6% |
| jayad | 15.1% |

TABLE IV
DISTRIBUTION OF QUESTION-ANSWER PAIRS AMONG SEASONS

| Query type | percentage |
|---|---|
| weather | 64.4% |
| plant protection | 17.8% |
| government schemes | 4.5% |
| nutrient management | 4.1% |
| cultural practises | 3.6% |
| fertilizers use | 2.9% |
| variety | 2.8% |

TABLE V
QUERY DISTRIBUTION AMONG QUERY TYPE

| What is the weather | 79.4% |
|---|---|
| How to control zinc deficiency in wheat | 2.3% |
| Information regarding macro nutrient management | 1.76% |

TABLE VI
MOST COMMON QUERIES

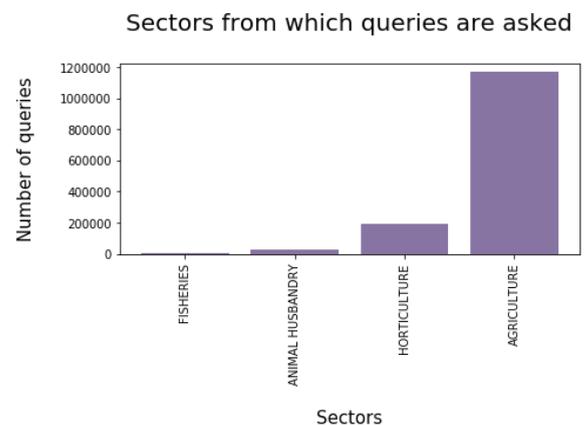

Fig. 1. Number of queries per sector

This analysis shows that only a limited number of unique queries are encountered while most of the queries are redundant. It also shows that the number of queries varies drastically from state to state. The answers to each query also differs on the basis of state and district from where the query has been asked.

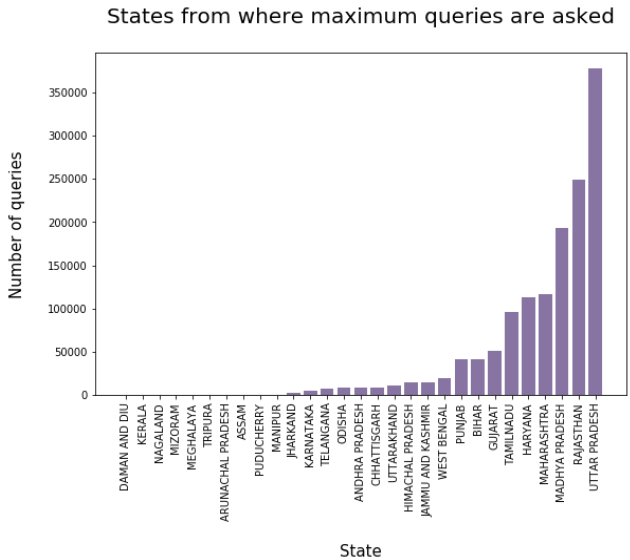

Fig. 2. Number of queries per state

## V. CHALLENGES

Apart from collecting the data, there were three significant challenges. First, we saw a lack of consistency in the format of the questions and answers. Most of the data is poorly written with many redundant words, spelling errors, incorrect grammar and punctuation. These features make the process of information extraction from the data a difficult task. The questions are well written compared to the answers in terms of the ease of understanding. Hence, we chose to process the questions to find the critical words. Also, the answers are very vague and are not framed sentences. Various answers are just numbers. Processing answers to understand their meaning and relevance to a particular input question is a challenging task. The given question is an example of a typical query in the data.

```
Q - caterpiler on grem damage?
A - spray quinolphos 30ml/15 l water
```

Second, some of the queries are registered in the regional language which poses a problem in pre-processing the data because the translation resources for specific languages are limited. Various questions and answers uses a few or all words from a native language. Also, most of them are not proper sentences. Such quality of data makes it difficult to process them even after translation to English.

Third, in order to check the accuracy of our system, we need a dataset with ground truth corresponding to each query which does not exist. Lack of truth values made it difficult to determine if the answer given as output for a given input query to the system is correct or not. Such a metric is necessary to measure the reliability of our model. Hence, the determination of a suitable metric for the model was a significant task.

## VI. METHODOLOGY

One of the crucial aspects of our model is sentence embedding which is done by using Sen2Vec model given by Arora et al.[2] We give a brief overview of the Sen2Vec model here, and the complete architecture is described in Figure 3.

*Sen2Vec Model*

The Sen2Vec model can be described as a method of converting a sentence into a vector, where the allotted weight to each dimension of the vector represents its inclination towards a particular context. The primary purpose of this model is to cluster the similar sentences without taking into consideration the ordering of the words.

Considering the improper format of the queries, we attempt to match input question to questions which are present in our given dataset rather than processing the answers - the idea being that given the size of the dataset and redundancy, the question is highly likely to be already present. We divided the collected data into two parts - train and test. Using the training data, we train our model based on Sen2Vec[2] and then for each query in test data we find the most similar question indexed in the training data.

### A. Pre-processing

For now, we consider English as a primary language and remove the queries registered in other languages. We then cleaned our data, which includes lower casing words, removing stop-words and stem to the roots. Then apart from using the current spell correct[11] for English dictionary, we develop our spell correct for local language words which may have appeared in the query mainly the crop names (Fig. 3). We then removed weather-related queries from our training data which will be compensated by a real-time weather API. We also remove redundancy using synonyms (Fig. 3). We looked at the frequency distribution of the words used in the queries and clubbed the queries if they use words with similar semantics. Finally, we group all answers for a particular question into a list to remove redundancy. Finally, the data-frame (containing the query, query-type, state, district, time of query and the list of answers corresponding to that query) is given as an input into our model.

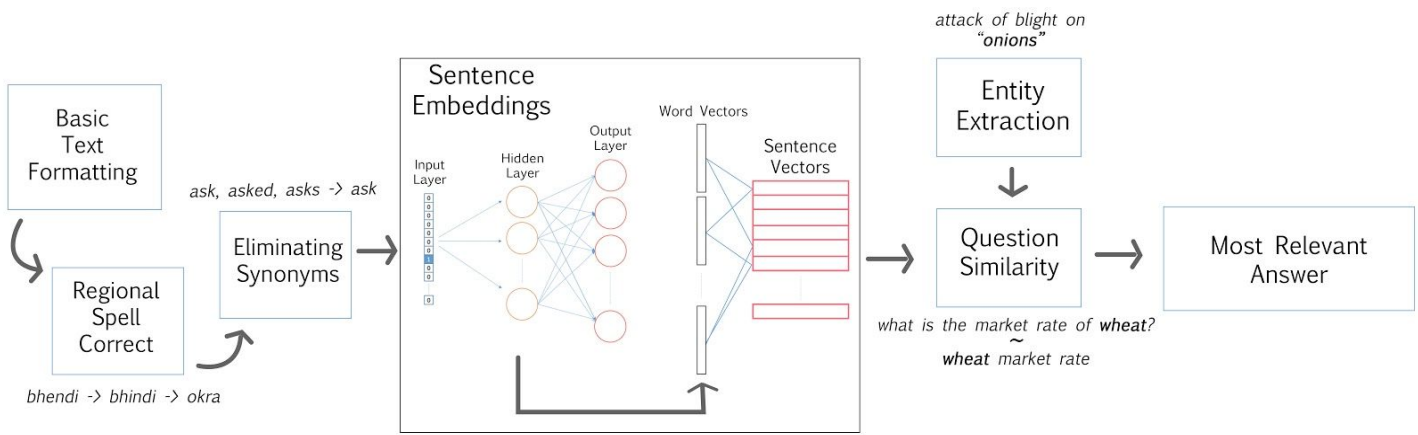

Fig. 3. Overview of the Sen2Vec model

B. *Training the model*

Given the pre-processed data-frame, we separated the data-frame into the train (80%) and test (20%). We train the word2Vec model[1]. We choose the model to train on 75 dimensions. The model outputs trained word embedding along with the required weights for the matrix for our model. These embeddings are converted to sentence embedding using the method described by Arora et al.[2]

C. *Embedding Optimization*

Now, there is an great chance to find questions which are highly similar however have different crop names for example, 'market rate of wheat?' and 'market rate of paddy?'. Since we know that the crop name is an essential determiner for the answer, we gave it a higher weight compared to other words. This was done by building an Entity Extractor which can be used to tag nouns[4] and then filter the crops from it. Using the entity extractor, we can identify crop names and give them higher weights (Fig. 3).

D. *Prediction*

We pre-processed the input query from the test dataset similarly and convert it into a vector using the embedding of the trained model. The model outputs the most similar query from the training data by comparing the embedding vectors using cosine similarity (Fig. 3). By similarity techniques, we obtain a list of answers which satisfy the input question. We applied an answer ranking method to output the best answer. The answer ranking takes input query calculates the lesk score with each answer from the list and output the one with the highest score.

VII. METRIC

For the metric, we wanted to capture similarity between the input sentence to the model and output predicted sentence from the model and use this to determine whether the two sentences are same.

We found that none of the standard scientific metrics to be suitable for evaluating our model. Because of the improper and inconsistent structure of question-answer pairs regarding language usage, we had to design a metric from scratch. Taking inspiration from Jaccard and Lesk similarity[3] metrics, we devised two metrics - modified Jaccard and modified Lesk scores in order to evaluate our model.

Our metric can be thought of as the amount of similarity between two sentences - input and prediction. Thus being able to find this value should give us a direct understanding of how our model is performing.

A. *Modified Jaccard Score*

We define our modified Jaccard score as the number of words in the intersection of the given question (known sentence) and the predicted question (predicted sentence), i.e.

$$Jaccard = \frac{count(knownSent \cap predictedSent)}{count(knownSent) + 1}$$

In this method, we simply use the words in the sentences as our parameters.

B. *Modified Lesk Score*

We first used words from meanings for various senses of words to create a gloss bag of words. We define our metric as the number of common words in the gloss bag of input question (known sentence) and the predicted question (predicted sentence) divided by the number of words in the gloss bag of the input question (known sentence), i.e.,

$$Lesk = \frac{count(gloss(knownSent) \cap gloss(predictedSent))}{count(gloss(knownSent)) + 1}$$

In both metrics we add 1 to the denominator to avoid zero division.

*Evaluating the metric*

In order to evaluate our metric, we manually labeled 100 test data queries and calculated our modified Jaccard scores and modified Lesk scores for the prediction of the test data questions. Using these predictions and the ground truth we then define a threshold for both scores. The threshold tells the model which predictions are to be considered as good results. We accordingly use the metrics for ranking our answers, where the final predicted answer is given by:

output answer = argmax[score(question, answer$_i$)]

VIII. RESULTS

Using the Modified Lesk score metric, our model was able to obtain an accuracy of about 56% without synonym elimination and entity extraction.

One key observation was that the crop names were important determiner while comparing the most similar queries. We thus performed entity extraction for the crop names. We had observed that the accuracy jumped from 56% to 86% after using entity extraction.

We then varied the dimension to improve accuracy. As demonstrated by the Fig. 4, the best performance of the model was observed at 75 number of dimensions for the embedding.

|         | Jaccard | Modified Lesk |
|---------|---------|---------------|
| Top - 1 | 64%     | 86%           |
| Top - 3 | 69%     | 89%           |
| Top - 5 | 70%     | 91%           |

TABLE VII
METRIC SCORE COMPARISON IN TOP-N MOST SIMILAR OUTPUT QUERIES

The table VII shows the metric scores for predictions when we trained our model for 75 dimension with synonym elimination and entity extraction for crop names.

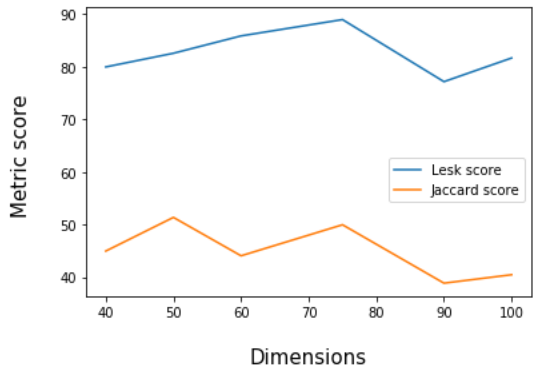

Fig. 4. Variation of the average metric score for the test data over the number of dimensions of the sentence embedding

Fig. 5. Example 1

$Q_{input}$ − what is the market rate of wheat?
$Q_{predicted}$ − wheat market rate

$A$ − wheat market rate − 1800 − − − 2200 rups pq

Fig. 6. Example 2

$Q_{input}$ − what is the fertilizer on grape?
$Q_{predicted}$ − fertilizer for grape

$A$ − spray saaf $20gm/15$ Litters of water for controlling blight on tomato

We also note that our chatbot can only answer pre-existing questions in the database. Absolutely new questions cannot be answered by our system. We plan to re-route such queries to human employees for answers. The newly created question-answer pair can be then added to our dataset for future reference to make the system adaptive to new queries.

## IX. CONCLUSION

Our chatbot can positively impact underserved communities by solving queries related to agriculture, horticulture and animal husbandry using natural language technology. The farmer will be able to receive agricultural information as well as localized information such as the current market prices of various crops in his/her district and weather forecast through an messaging app. A farmer can directly message our AI enabled the system in his/her language, and get an answer. Our system would enable the farmer to ask any number of questions, anytime, which will in turn help in spreading the modern farming technology faster and to a higher number of farmers.

Moreover, we found that most of the queries related to localized information such as weather and market prices were redundant. Our Question-Answer system can answer maximum queries on its own without any human intervention with high accuracy. This will lead to better utilization of human resource and avoid unnecessary costs in setting up new call centers. Our system is capable of handling all the redundant queries and getting updated with new queries on the go. The system also provides an option that enables the farmer to ask questions directly to the KCC employees if and when necessary.

Above all, we believe that the system helps in analyzing the farmers' mindset and the structure of the Agricultural Sector in India. While the system provides a secure communication channel to the farmer, it also helps the policy makers to understand the needs and concerns of the farmers. The data analysis also provides an understanding of which sector or season farmers requires attention. Thus, our decision support system uses all the available resources judiciously to tackle the problem of lack of awareness and information in the agricultural sector in India.

## X. FUTURE WORK

For the future, we plan to improve the answer ranking mechanism, implement multilingual support for the chatbot with voice-over support and entity extraction from answers for generating knowledge graphs.